\documentclass[10pt,draftcls,journal,compsoc,letterpaper,oneside,onecolumn]{bib/IEEEtran}
\ifx\pdfoutput\undefined
  \usepackage[dvipdfm]{graphicx}  
  \usepackage[dvipdfm,bookmarks=false]{hyperref}
\else
  \usepackage[pdftex]{graphicx}  
  \usepackage[pdftex,bookmarks=false]{hyperref}
\fi
\usepackage[fleqn]{amsmath}
\usepackage[psamsfonts]{amssymb}

\newtheorem{hosoku}{Remark}

\def\u{\boldsymbol{u}}

\def\x{\boldsymbol{x}}
\def\y{\boldsymbol{y}}
\def\z{\boldsymbol{z}}

\def\X{\boldsymbol{X}}

\def\0{\boldsymbol{0}}

\def\cL{\mathcal L}

\def\cN{\mathcal N}
\def\cO{\mathcal O}

\def\ov{\overline}
\def\ul{\underline}
\def\wh{\widehat}
\def\wt{\widetilde}

\begin{document}

%
\title{%
  A stochastic model of human visual attention with a dynamic Bayesian network
}
\author{%
  Akisato Kimura,~\IEEEmembership{Senior Member,~IEEE,}
  Derek Pang,~\IEEEmembership{Student Member,~IEEE,}
  Tatsuto Takeuchi,~\IEEEmembership{}
  Kouji Miyazato,~\IEEEmembership{}
  Kunio Kashino,~\IEEEmembership{Senior Member,~IEEE,} and
  Junji Yamato,~\IEEEmembership{Senior Member,~IEEE.}
  \IEEEcompsocitemizethanks{%
    \IEEEcompsocthanksitem
    The authors are with NTT Communication Science Laboratories, NTT Corporation, 3-1
    Morinosato Wakamiya, Atsugi, Kanagawa, 243-0198 Japan.
    E-mail: akisato@ieee.org
    \IEEEcompsocthanksitem
    D. Pang is with Department of Electrical Engineering, Stanford University, Packard
    240, 350 Serra Mall, Stanford, CA 94305, USA. He contributed to this work during his
    internship at NTT Communication Science Laboratories.
    \IEEEcompsocthanksitem
    K. Miyazato was with Department of Information and Communication Systems Engineering,
    Okinawa National College of Technology, 905 Henoko, Nago, Okinawa, 905-2192 Japan.
    He contributed to this work during his internship at NTT Communication Science
    Laboratories.
    \IEEEcompsocthanksitem
    Parts of the material in this paper has been presented at IEEE International
    Conference on Multimedia and Expo (ICME2008), Hannover, Germany, June 2008, and IEEE
    International Conference on Multimedia and Expo (ICME2009), Cancun, Mexico, June-July
    2009.
    \IEEEcompsocthanksitem
    Manuscript receive March 31 2010.
  }
}
\markboth{IEEE TRANSACTIONS ON PATTERN ANALYSIS AND MACHINE INTELLIGENCE,~Vol.~xxx, No.~xxx,~xxxxx~2010}{Kimura \MakeLowercase{\textit{et al.}}: A STOCHASTIC MODEL OF HUMAN VISUAL ATTENTION WITH A DYNAMIC BAYESIAN NETWORK}

\allowdisplaybreaks  

\maketitle

\begin{abstract}
Recent studies in the field of human vision science suggest that the human responses to
the stimuli on a visual display are non-deterministic. People may attend to different
locations on the same visual input at the same time. Based on this knowledge, we propose
a new stochastic model of visual attention by introducing a dynamic Bayesian network to
predict the likelihood of where humans typically focus on a video scene. The proposed
model is composed of a dynamic Bayesian network with 4 layers. Our model provides a
framework that simulates and combines the visual saliency response and the cognitive
state of a person to estimate the most probable attended regions. Sample-based inference
with Markov chain Monte-Carlo based particle filter and stream processing with multi-core
processors enable us to estimate human visual attention in near real time. Experimental
results have demonstrated that our model performs significantly better in predicting
human visual attention compared to the previous deterministic models.
\end{abstract}


\section{Introduction}
\label{sec:intro}

Developing a sophisticated object detection and recognition algorithms has been a long
distance challenge in computer and robot vision researches. Such algorithms are required
in most applications of computational vision, including robotics \cite{roboticsSaliency:%
asada}, medical imaging \cite{medicalImagingSaliency}, intelligent cars
\cite{intelligentCarsSaliency}, surveillance \cite{nonparaBayesAttention}, image
segmentation \cite{foveatedShotDetection:boccignone,salientRegionExtraction:fukuchi_ICME}
and content-based image retrieval \cite{nearIdenticalShotDetection:zisserman}. One of the
major challenges in designing generic object detection and recognition systems is to
construct methods that are fast and capable of operating on standard computer platforms
without any prior knowledge. To that end, pre-selection mechanism would be essential to
enable subsequent processing to focus only on relevant data. One promising approach to
achieve this mechanism is visual attention: it selects regions in a visual scene that are
most likely to contain objects of interest. The field of visual attention is currently
the focus of much research for both biological and artificial systems.

\begin{figure}[t]
  \begin{center}
    \includegraphics[width=0.985\hsize,keepaspectratio]{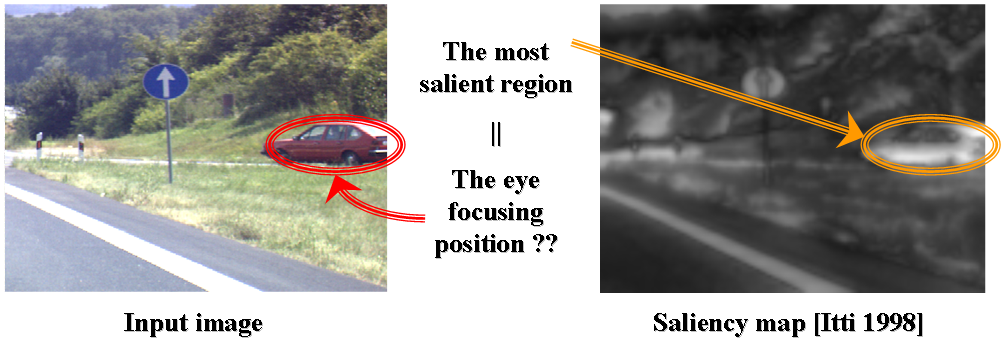}
    \caption{An example of a saliency map using Koch-Ullman model}
    \label{fig:saliencymap}
  \end{center}
\end{figure}

Attention is generally controlled by one or a combination of the two mechanisms: 1) a
top-down control that voluntarily chooses the focus of attention in a cognitive and
task-dependent manner, and 2) a bottom-up control that reflexively directs the visual
focus based on the observed saliency attributes. The first biologically-plausible model
for explaining the human attention system was proposed by Koch and Ullman
\cite{visualAttention:KochUllman}, which follows the latter approach. The basic concept
underlying this model is the {\it feature integration theory} developed by Treisman and
Gelade \cite{FeatureIntegrationTheory:Treisman} which has been one of the most
influential theories of human visual attention. According to the feature integration
theory, in a first step to visual processing, several primary visual features are
processed and represented with separate {\it feature maps} that are later integrated in a
{\it saliency map} that can be accessed in order to direct attention to the most
conspicuous areas. In an example shown in Fig. \ref{fig:saliencymap}, a red car placed on
the right in the frame should be attentive, and therefore people directs one's attention
to this area. The Koch-Ullman model has been attracting attention of many researchers,
especially after the development of an implementation model by Itti, Koch and Niebur
\cite{Saliency:itti}. Later, so many attempts have been made to improve the Koch-Ullman
model \cite{VisualAttention:Privitera,GeneratingEyeFixations:gu,VOCUS,saliency:minho2,%
decisionTheorySaliency:gao} and to extend it to video signals \cite{decisionTheory%
Saliency:gao,DetectSurprise:itti,videoSaliency:clement,saliency:minho1}.

\begin{figure}[t]
  \begin{center}
    \includegraphics[width=0.985\hsize,keepaspectratio]{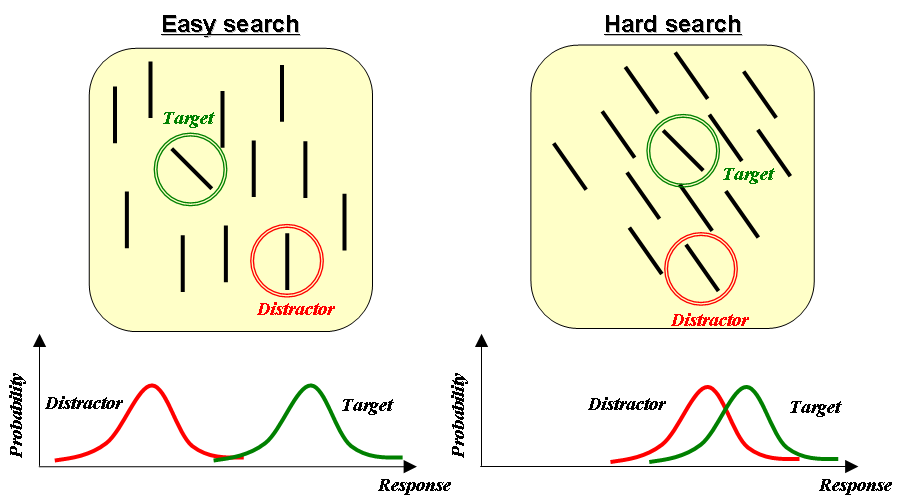}
    \caption{
      Visual search response based on the signal detection theory proposed by
      Eckstein et al.
      \cite{SignalDetectionTheory:Eckstein,SignalDetectionTheory:Verghese}.
    }
    \label{fig:sdt}
  \end{center}
\end{figure}

Although the feature integration theory well explains the early human visual system,
a part of the theory includes one crucial problem, namely, people may attend to different
locations on the same visual input at the same time. The example shown in Fig. \ref{fig:%
saliencymap} exactly indicates the phenomena: people may pay attention to a blue traffic
sign at the center, a white line at the bottom left or others. Previously, this
inconsistent visual attention has been considered to be caused by object-based attention,
rather than location-based attention \cite{objectsAttention:Scholl2002}, which implies
that inconsistent visual attention are heavily controlled by higher order processes such
as top-down intention, knowledge and preferences. Another
typical example can be seen in Fig. \ref{fig:sdt}. Let us consider a search task with a
single $45^{\circ}$ target among a lot of distractors. We can easily understand that the
left case is easy and the right case is difficult to find the target. However, based on
the feature integration theory, we can immediately identify the target for both easy and
hard searches since we always select the location where the response of the detector
tuned to the target visual property is greater than at any other locations.

On the other hand, another theory to understanding visual search and attention has been of
interest, called the \textit{signal detection theory} \cite{SignalDetectionTheory:%
Eckstein,SignalDetectionTheory:Verghese}. According to this theory, the elements in a
visual display are internally represented as independent random variables. Again let us
consider the search task shown in Fig. \ref{fig:sdt}. The response of a detector tuned to
the target orientation is represented as a Gaussian density. The response of the same
detector to the distractor is also a Gaussian density with lower mean value. For a
$45^{\circ}$ target and vertical distractors, these densities barely overlap, which
implies that we can immediately detect the target. On the other hand, in the case of hard
search, the target density is identical to the easy search case, but the distractor
density is shifted rightward, so that the two densities corresponding to the target and
distractor overlap. This implies that the probability we focus on the distractors becomes
high and therefore it takes much time to detect the target.

With the paradigm of the signal detection theory, we proposes a new stochastic model of
visual attention. With this model, we can automatically predict the likelihood of where
humans typically focus on a visual input. The proposed model is composed of a dynamic
Bayesian network with four layers: (1) \textit{a saliency map} that shows the average
saliency response at each position of a video frame, (2) \textit{a stochastic saliency
map} that converts the saliency map into a natural human response through a Gaussian
state space model based on the finding of the signal detection theory, 3) an \textit{eye
movement pattern} that controls the degree of ``overt shifts of attentions'' (shifts with
saccadic eye movements) through a hidden Markov model (HMM), and  4) an \textit{eye focusing
density map} that predicts positions that people probably pay attention to based on the
stochastic saliency map and eye movement patterns. When describing the Bayesian network
of visual attention, the principle of the signal detection theory is introduced, namely,
the position where values of the stochastic saliency map takes the maximum is the eye
focusing positions. The proposed model also provides a framework simulating top-down
cognitive states of a person at the layer of eye movement patterns. The introduction of
eye movement patterns as hidden states of HMMs enables us to describe the mechanism of
eye focusing and eye movement naturally.

The paper is organized as follows: Section \ref{sec:related} discusses several related
researches that focuses on modeling of human visual attention by using probabilistic
techniques or concepts. Section \ref{sec:model} describes the proposed stochastic model
of visual attention. Section \ref{sec:parameter} presents the methods for finding maximum
likelihood (ML) estimates of the model parameters based on the Expectation-Maximization
(EM) framework. Section \ref{sec:eval} discusses some evaluation results. Finally Section
\ref{sec:conclude} summarizes the report and discusses future work.

\section{Related work}
\label{sec:related}

Several previous researches focused on modeling of human visual attention by using some
kind of probabilistic techniques or concepts. Itti and Baldi \cite{DetectSurprise:itti}
investigated a Bayesian approach to detecting surprising events in video signals. Their
approach models a surprise by Kullback-Leibler divergence between the prior and posterior
distributions of fundamental features. Avraham and Lindenbaum \cite{AvrahamLindenbaum%
Paper} utilized a graphical model approximation to extend their static
saliency model based on self similarities. Boccignone \cite{nonparaBayesAttention}
introduced a nonparametric Bayesian framework to achieve object-based visual attention.
Gao, Mahadevan and Vasconcelos \cite{decisionTheorySaliency:gao,spaciotemporalSaliency}
developed a decision-theoretic approach attention model for object detection.

The main contribution of our stochastic model against the above previous researches is
the introduction of a unified stochastic model that integrates ``covert shifts of
attention'' (shifts of attentions without saccadic eye movements) driven by bottom-up
saliency with ``overt shifts of attention'' (shifts of attention with saccadic eye
movements) driven by eye movement patterns by using a dynamic Bayesian network. Our
proposed model also provides a framework that simulates and combines the bottom-up visual
saliency response and the top-down cognitive state of a person to estimate probable
attended regions, if eye movement patterns can deal with more sophisticated top-down
information. How to integrate such kinds of top-down information is one of the most
important future researches.

\begin{figure}[t]
  \begin{center}
    \includegraphics[width=0.975\hsize,keepaspectratio]{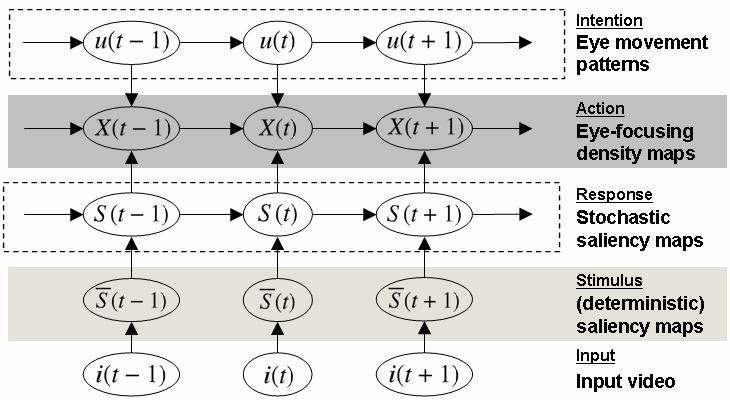}
    \caption{
      Graphical representation of the proposed stochastic model of human visual
      attention, where arrows indicate stochastic dependencies.
    }
    \label{fig:graphicalModel}
  \end{center}
\end{figure}

\begin{figure*}[t]
  \begin{center}
    \includegraphics[width=0.975\hsize,keepaspectratio]{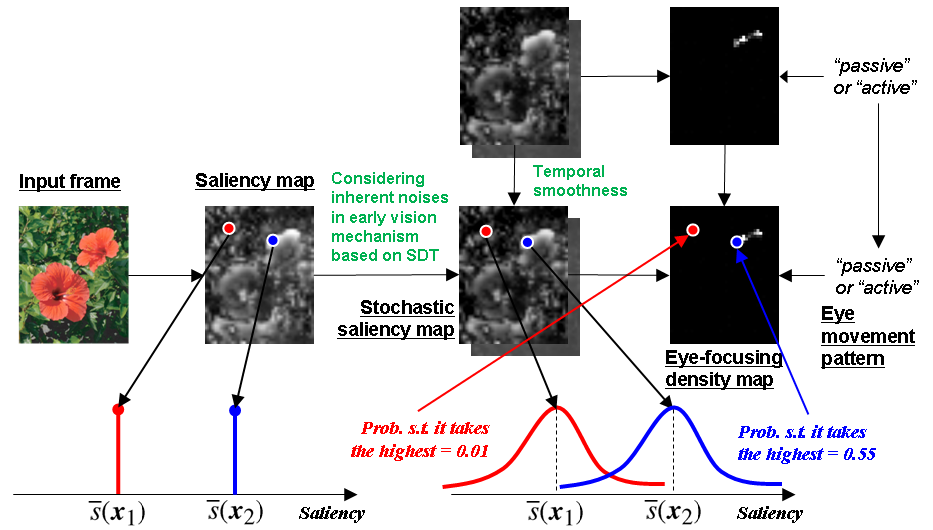}
    \caption{Procedure of the proposed model}
    \label{fig:graphicalModel2}
  \end{center}
\end{figure*}

\section{Stochastic visual attention model}
\label{sec:model}

\subsection{Overview}
\label{sec:model:overview}

Figs. \ref{fig:graphicalModel} and \ref{fig:graphicalModel2} illustrates the graphical
representation of the proposed visual attention model. The proposed model consists of
four layers: (deterministic) saliency maps, stochastic saliency maps, eye focusing
positions and eye movement patterns. Before describing the model of the proposed visual
attention model, let us introduce several notations and definitions.

$I=\allowbreak i(1:T)=\allowbreak\{i(t)\}_{t=1}^T$ denotes an input video, where $i(t)$
is the $t$-th frame of the video $I$ and $T$ is the duration (i.e. the total number of
frames) of the video $I$. The symbol $I$ also denotes a set of coordinates in the frame.
For example, a position $\y$ in a frame is represented as $\y\in I$.

$\ov{S}=\allowbreak \ov{S}(1:T)=\allowbreak \{\ov{S}(t)\}_{t=1}^T$ denotes a
\textit{saliency video} which comprises a sequence of \textit{saliency maps}
$\ov{S}(t)$ obtained from the input video $I$. Each saliency map is denoted as
$\ov{S}(t)=\allowbreak\{\ov{s}(t,\y)\}_{\y\in I}$, where $\ov{s}(t,\y)$ is
called \textit{saliency} which is the pixel value at the position $\y\in I$. Each
saliency represents the strength of visual stimulus on the corresponding position of a
frame with the real value between 0 and 1.

$S=\allowbreak S(1:T)=\allowbreak\{S(t)\}_{t=1}^T$ denotes a \textit{stochastic
saliency video} which comprises a sequence of \textit{stochastic saliency maps} $S(t)$
obtained from the input video $I$. Each stochastic saliency map is denoted as
$S(t)=\allowbreak\{s(t,\y)\}_{\y\in I}$, where
$s(t,\y)$ is called \textit{stochastic saliency} which is the pixel value at the
position $\y\in I$. Each stochastic saliency corresponds to saliency response perceived
through a certain kind of random processes.

$U=\allowbreak u(1:T)=\allowbreak\{u(t)\}_{t=1}^T$ denotes a sequence of
\textit{eye movement patterns} each of which represents a pattern of eye movements, A
previous research \cite{beyondBottomUp} implies that there are two typical patterns
\footnote{Peters and Itti \cite{beyondBottomUp} prepared the other pattern, interactive
state, which can be seen when playing video games, driving a car or browsing webs. We will
omit the interactive state since our setting in this paper does not include any
interactions} of
eye movements when one is simply watching a video: 1) Passive state $u(t)=0$ in which
one tends to stay around one particular position to continuously capture important visual
information, and 2) active state $u(t)=1$ in which one actively moves around and searches
various visual information on the scene. Eye movement patterns may reflect purposes or
intentions of human eye movements.

$X=\allowbreak X(1:T)=\allowbreak\{\x(t)\}_{t=1}^T$ denotes a sequence of eye
focusing positions. The proposed model estimates the eye focusing position by integrating
the bottom-up information (stochastic saliency maps) and the top-down information (eye
movement patterns). A map that represents a density of eye focusing positions is called
an \textit{eye focusing density map}.

Only the saliency maps are observed, and therefore eye focusing positions should be
estimated under the situation where other layers (stochastic saliency maps and eye
movement patterns) are hidden.

In what follows, we denote a probability density function (PDF) of an $x$ as
$p(x)$, a conditional PDF of an $x$ given $y$ as $p(x|y)$, and a PDF of $x$ with
a parameter $\theta$ as $p(x;\theta)$.

The rest of this section describes the detail of the proposed stochastic model and the
method for estimating eye focusing positions only from input videos.

\subsection{Saliency maps}
\label{sec:model:saliency}

\begin{figure}[t]
  \begin{center}
    \includegraphics[width=0.985\hsize,keepaspectratio]{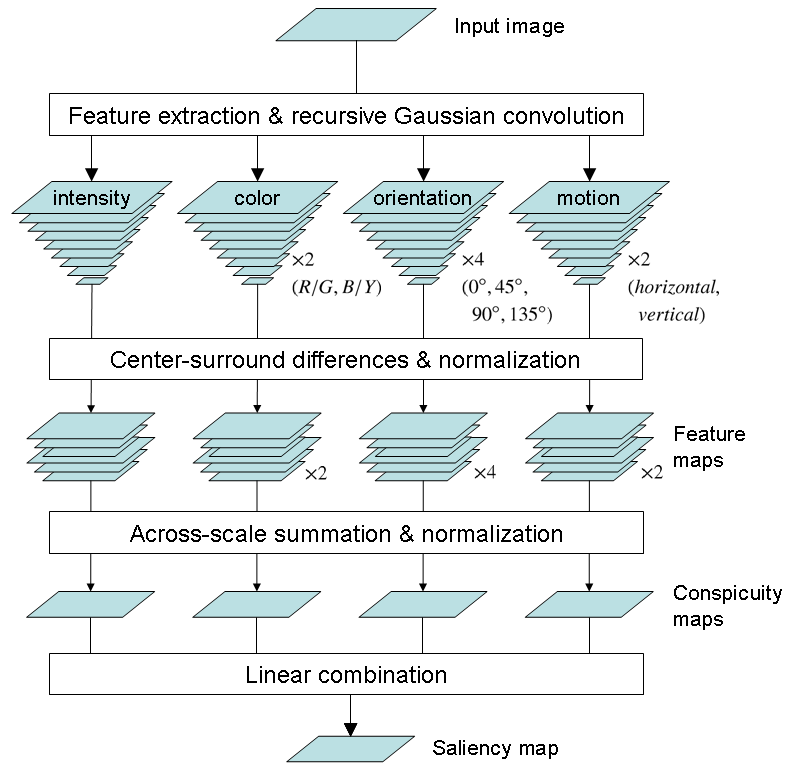}
    \caption{Saliency map extraction using Itti et al.'s model}
    \label{fig:itti}
  \end{center}
\end{figure}

We used Itti-Koch saliency model \cite{Saliency:itti} shown in Fig. \ref{fig:itti} to
extract (deterministic) saliency maps. Our implementation includes twelve feature
channels sensitive to color contrast (red/green and blue/yellow), temporal luminance
flicker, luminance contrast, four orientations ($0^{\circ}$, $45^{\circ}$, $90^{\circ}$
and $135^{\circ}$), and two oriented motion energies (horizontal and vertical). These
features detect spatial outliers in image space using a center-surround architecture.
Center and surround scales are obtained from dyadic pyramids with 9 scales, from scale 0
(the original image) to scale 8 (the image reduced by a factor of $2^8 = 256$ in both the
horizontal and vertical dimensions). Six center-surround difference maps are then
computed as point-wise differences across pyramid scales, for combinations of three
center scales ($c = \{2,3,4\}$) and two center-surround scale differences ($s=\{3,4\}$).
Each feature map is additionally endowed with internal dynamics that provide a strong
spatial within-feature and within-scale competition for activity, followed by
within-feature, across-scale competition. In this way, initially noisy feature maps can
be reduced to sparse representations of only outlier locations which stand out from their
surroundings. All feature maps finally contribute to a unique saliency map representing
the conspicuity of each location in the visual field. The saliency map is adjusted with a
centrally-weighted 'retinal' filter, putting a higher emphasizes on the saliency values
around the center of the video.

\subsection{Stochastic saliency maps}
\label{sec:model:stochastic}

When estimating a \textit{stochastic saliency map} $S(t)=\{s(t,\y)\}_{\y\in I}$, we
introduce a pixel-wise state space model characterized by the following two
relationships:
\begin{eqnarray*}
  p(\ov{s}(t,\y)|s(t,\y)) &=& \cN(s(t,\y),\sigma_{s1}),\\
  p(s(t,\y)|s(t-1,\y)) &=& \cN(s(t-1,\y),\sigma_{s2}),
\end{eqnarray*}
where $\cN(\ov{s},\sigma)$ is the Gaussian PDF with mean $\ov{s}$ and variance
$\sigma^2$. The first equation in the above model implies that a saliency map is observed
through a Gaussian random process, and the second equation exploits the temporal
characteristics of the human visual system.  For brevity, only in this section we will
omit the position $\y$ where explicit expression is unnecessary, e.g. $s(t)$ instead of
$s(t,\y)$.

We employ a Kalman filter to recursively compute the stochastic saliency map.
Assume that the density at each position on the stochastic saliency map $s(t-1)$ at
time $t-1$ given saliency maps $\ov{s}(1:t-1)$ up to time $t-1$ is given as the following
Gaussian PDF:
\begin{eqnarray*}
  \lefteqn{p(s(t-1)|\ov{s}(1:t-1))}\\
  &=& \cN(\wh{s}(t-1|t-1),\sigma_s(t-1|t-1)).
\end{eqnarray*}
where the position $\y$ is omitted for simplicity. Then, the density
$p(s(t)|\ov{s}(1:t))$ of the stochastic saliency map at time $t$ is updated by the
following recurrence relations with the saliency maps $\ov{s}(1:t)$ up to time $t$:

\medskip\noindent\textbf{[Estimation step]}
\begin{eqnarray*}
  p(s(t)|\ov{s}(1:t-1)) &=& \cN(\wh{s}(t|t-1),\sigma_s(t|t-1)),
\end{eqnarray*}
where
\begin{eqnarray*}
  \wh{s}(t|t-1) &=& \wh{s}(t-1|t-1),\\
  \sigma_s^2(t|t-1) &=& \sigma_{s2}^2+\sigma_s^2(t-1|t-1).
\end{eqnarray*}
\noindent\textbf{[Update step]}
\begin{eqnarray*}
  p(s(t)|\ov{s}(1:t)) &=& \cN(\wh{s}(t|t),\sigma_s(t|t)),
\end{eqnarray*}
where
\begin{eqnarray}
  \lefteqn{\wh{s}(t|t) =} \label{eq:model:stochastic:kalman1}\\
  && \frac{\sigma_{s1}^2}{\sigma_{s1}^2+\sigma_s^2(t|t-1)}\wh{s}(t|t-1)+
     \frac{\sigma_s^2(t|t-1)}{\sigma_{s1}^2+\sigma_s^2(t|t-1)}\ov{s}(t),\nonumber\\
  && \sigma_s^2(t|t)
     =\frac{\sigma_{s1}^2\cdot\sigma_s^2(t|t-1)}{\sigma_{s1}^2+\sigma_s^2(t|t-1)},
      \label{eq:model:stochastic:kalman2}
\end{eqnarray}

\begin{hosoku}\label{rem:online_params}
The above model implies that model parameters ($\sigma_{s1}$, $\sigma_{s2}$) of every
Gaussian random variable is independent from the frame index $t$ and the position $\y$.
We can easily extend the model to consider adaptive model parameters depending on the
frame index and the position. In this case, model parameters can be updated via on-line
learning with adaptive Kalman filters (e.g. \cite{adaptiveKalman:myers,adaptiveKalman:%
leathrum}.)\quad\textit{(Remark\ \ref{rem:online_params} ends.)}
\end{hosoku}

\subsection{Estimating eye motions}
\label{sec:model:eyemotion}

\subsubsection{Overview}
\label{sec:model:eyemotion:overview}

By incorporating the stochastic saliency map $S(t)=\{s(t,\y)\}_{\y\in I}$ and the
\textit{eye movement pattern} $u(t)$, we introduce the following transition PDF to
estimate the \textit{eye focusing position} $\x(t)$ such that
\begin{eqnarray}
  \lefteqn{p(\x(t),u(t)|p(S(t)),\x(t-1),u(t-1))}\nonumber\\
  &\propto& p(\x(t)|p(S(t)))\nonumber\\
  &       & \cdot p(u(t)|u(t-1))\cdot p(\x(t)|\x(t-1),u(t)),
            \label{eq:model:eyemotion:prob4}
\end{eqnarray}
where the PDF of the stochastic saliency map at time $t$ is represented as $p(S(t))$ for
simplicity, namely
\begin{eqnarray*}
  p(S(t))    &=& \{p(s(t,\y))\}_{\y\in I},\\
  p(s(t,\y)) &=& p(s(t,\y)|\ov{s}(1:t,\y)) \quad\forall\y\in I.
\end{eqnarray*}
The stochastic saliency map $S(t)$ controls ``covert shifts of attention'' through the
PDF $p(\x(t)|p(S(t)))$ \footnote{The notation $p(\x(t)|p(S(t)))$ seems to be unusual,
however, the PDF of eye focusing positions $\x(t)$ estimated from the stochastic saliency
map $S(t)$ can be determined by the PDF of the stochastic saliency map, not the
stochastic saliency map itself, as shown in Section \ref{sec:model:eyemotion:bottomup}.}.
On the other hand, the eye movement pattern $u(t)$ controls the degree of ``overt shifts
of attention''. In what follows, we call a pair $z(t)=(\x(t),u(t))$, consisting of an eye
focusing position and an eye movement pattern, as the \textit{eye focusing state} $z(t)$
for brevity. The following PDF of eye focusing positions $\x(t)$ given a PDF $p(S(1:t))$
of stochastic saliency maps up to time $t$ characterizes an eye focusing density map at
time $t$:
\begin{eqnarray}
  \lefteqn{p(\x(t)|p(S(1:t)))=\sum_{u(t)=0,1}p(z(t)|p(S(1:t))),}
    \label{eq:eyemotion:overview:1}\\
  \lefteqn{p(z(t)|p(S(1:t)))=\int_{z(t-1)}\hspace{-4mm}p(z(t-1)|p(S(1:t-1)))}\nonumber\\
  & & \hspace{15mm}\cdot p(z(t)|p(S(t)),z(t-1))dz(t-1).
    \label{eq:eyemotion:overview:2}
\end{eqnarray}

Since the formula for computing Eq. (\ref{eq:eyemotion:overview:1}) cannot be
derived, we introduce a technique inspired by a particle filter with Markov chain
Monte-Carlo (MCMC) sampling instead. The PDF of eye focusing states shown in Eq.
(\ref{eq:eyemotion:overview:2}) can be approximated by samples of eye focusing states
$\{z_n(t)\}_{n=1}^N$ and the associated weights $\{w_n(t)\}_{n=1}^N$ as
\begin{eqnarray}
  p(z(t)|p(S(1:t))) &\approx& \sum_{n=1}^N w_n(t)\cdot\delta(z(t),z_n(t)),
    \label{eq:eyemotion:overview:approx}
\end{eqnarray}
where $N$ is the number of samples, and $\delta(\cdot,\cdot)$ represents Kronecker delta.

\begin{figure}[t]
  \begin{center}
    \includegraphics[width=0.985\hsize,keepaspectratio]{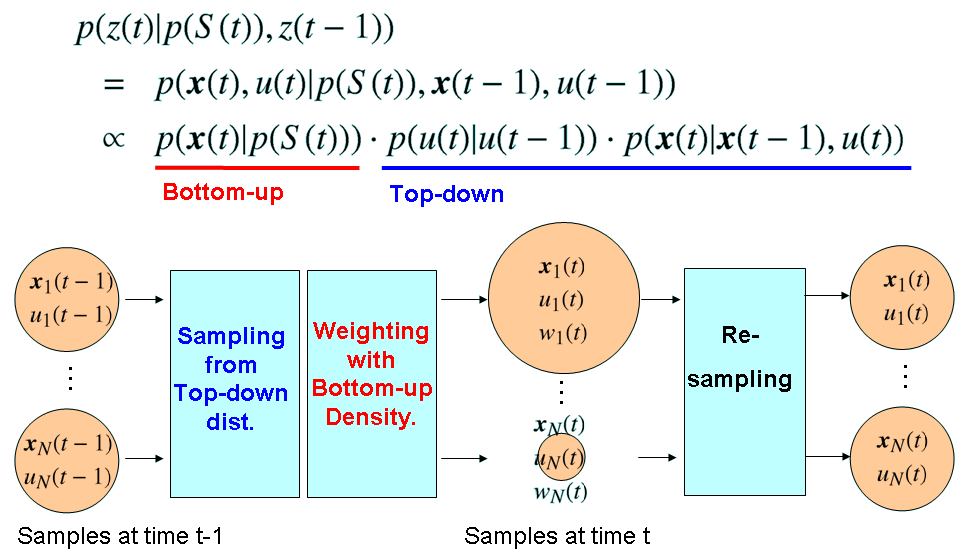}
    \caption{Strategy for calculating eye focusing density maps}
    \label{fig:efdm}
  \end{center}
\end{figure}

Fig. \ref{fig:efdm} shows the procedure for estimating eye focusing density maps, which
can be separated into three steps: 1) generating samples from the PDFs $p(u(t)|u(t-1))$
and $p(\x(t)|\x(t-1),u(t))$ derived from an eye movement pattern, 2) weighting samples
with the PDF $p(\x(t)|p(S(t)))$ derived from a stochastic saliency map, and 3) re-sampling
if necessary. We now describe each step in detail.

\subsubsection{Propagation with eye movement patterns}
\label{sec:model:eyemotion:topdown}

The second and third terms of Equation (\ref{eq:model:eyemotion:prob4}) suggests that
the current eye focusing position depends on the previous eye focusing position, and the
degree of eye movements is driven by one's eye movement pattern $u(t)$.

The second term $p(u(t)|u(t-1))$ of Equation (\ref{eq:model:eyemotion:prob4}) is
characterized by the the transitional probability $\Phi=\{\phi_{(i,j)}\}$ of eye movement
patterns defined by a $2\times 2$ matrix given in advance.
\begin{eqnarray}
  p(u(t)|u(t-1)) &=& \phi_{(u(t),u(t-1))} \label{eq:model:eyemotion:prob2}
\end{eqnarray}

\begin{figure}[t]
  \begin{center}
    \includegraphics[width=0.985\hsize,keepaspectratio]{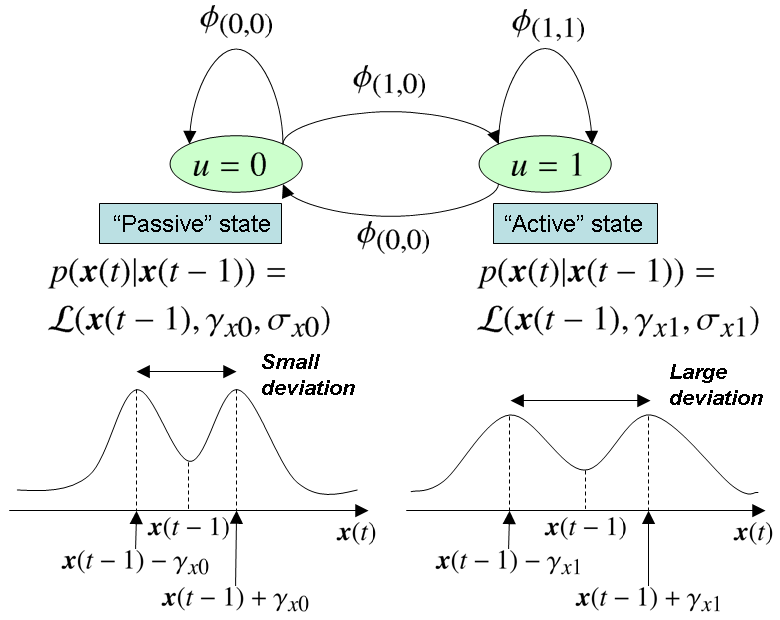}
    \caption{
      Transition PDF of eye focusing positions governed by the eye movement patterns
    }
    \label{fig:eyepositionHMM}
  \end{center}
\end{figure}

The third term $p(\x(t)|\x(t-1),u(t))$ of Equation (\ref{eq:model:eyemotion:prob4})
represents the transition PDF of eye focusing positions governed by the eye movement
patterns at the current time (See Figure \ref{fig:eyepositionHMM}), defined as
\begin{eqnarray}
  p(\x(t)|\x(t-1),u(t)) &=& \cL(\x(t-1),\gamma_{x,u(t)},\sigma_{x,u(t)}),
  \label{eq:model:eyemotion:prob3}
\end{eqnarray}
where $\gamma_{xi}$ and $\sigma_{xi}$ $(i=1,2)$ are model parameters that represents the
average and standard deviation of distances of eye movements, and $\cL(\ov{\x},\gamma,%
\sigma)$ is a shifted 2D Gaussian PDF with mean $\ov{\x}$, indent $\gamma$ and variance
$\sigma^2$ such that
\begin{eqnarray*}
  \cL(\ov{\x},\gamma,\sigma)
  &\propto& \exp\left\{-\frac{(\|\x-\ov{\x}\|-\gamma)^2}{2\sigma^2}\right\}.
\end{eqnarray*}

Samples $\{\z_n(t)\}_{n=1}^N$ of eye focusing states are generated with a technique of
MCMC sampling. Suppose that samples $\{\z_n(t-1)\}_{n=1}^N$ of eye focusing states at
time $t-1$ have already been obtained. Then, samples $\{\z_n(t)\}_{n=1}^N$ at time $t$
are drawn by using the second and third terms of Equation (\ref{eq:model:eyemotion:%
prob4}) with the Metropolis algorithm \cite{Metropolis} such as
\begin{eqnarray*}
  \z_n(t) &=& \{\x_n(t),\u_n(t)\},\\
  u_n(t)  &\sim& p(u(t)|u_n(t-1)),\\
  \x_n(t) &\sim& p(\x(t)|\x_n(t-1),u_n(t)),
\end{eqnarray*}
where $\x\sim p(\x)$ indicates that a sample $\x$ is drawn from a PDF $p(\x)$. This
top-down part corresponds to the propagation step of a particle filter.

\subsubsection{Updating with stochastic saliency maps}
\label{sec:model:eyemotion:bottomup}

As the second step, sample weights $\{w_n(t)\}_{n=1}^N$ are updated based on the first
term $p(\x(t)|p(S(t)))$ of Equation (\ref{eq:model:eyemotion:prob4}). Formally, the
weight $w_n(t)$ of the $n$-th sample $z_n(t)$ at time $t$ can be calculated as
\begin{eqnarray*}
  w_n(t) &\propto& w_n(t-1)\cdot p(\x(t)=\x_n(t)|p(S(t))).
\end{eqnarray*}
As shown in Equation (\ref{eq:eyemotion:overview:approx}), samples $\{\z_n(t)\}_{n=1}^N$
of eye focusing states and the associated weights $\{w_n(t)\}_{n=1}^N$ comprise
an eye focusing density map at time $t$. This step corresponds to the update step of a
particle filter.

The first term of Equation (\ref{eq:model:eyemotion:prob4}) represents the fact that the
eye focusing position is selected based on the signal detection theory, where the
position at which the stochastic saliency takes the maximum is determined as the eye
focusing position. In other words, this term computes the probability at each position
that the stochastic saliency takes the maximum, which can be calculated as
\begin{eqnarray}
  \lefteqn{p(\x(t)|p(S(t)))}\nonumber\\
  &=& \int_{-\infty}^{\infty}p(s(t,\x(t))=s)\prod_{{\wt{\x}\in I,}
      \atop{\wt{\x}\neq\x(t)}}P(s(t,\wt{\x})\le s)ds,\hspace{5mm}
      \label{eq:model:eyemotion:prob1}
\end{eqnarray}
where $P(s(t,\wt{\y})\le s)$ is the cumulative density function (CDF) that corresponds to
the PDF $p(s(t,\wt{\y}))$ of the stochastic saliency $s(t,\wt{\y})$. The first part of
Equation (\ref{eq:model:eyemotion:prob1}) stands for the probability such that a
stochastic saliency value at position $\x(t)$ equals $s$, and the second part represents
the probability such that stochastic saliency values at any other positions are smaller
than $s$.
 
\begin{figure}[t]
  \begin{center}
    \includegraphics[width=0.985\hsize,keepaspectratio]{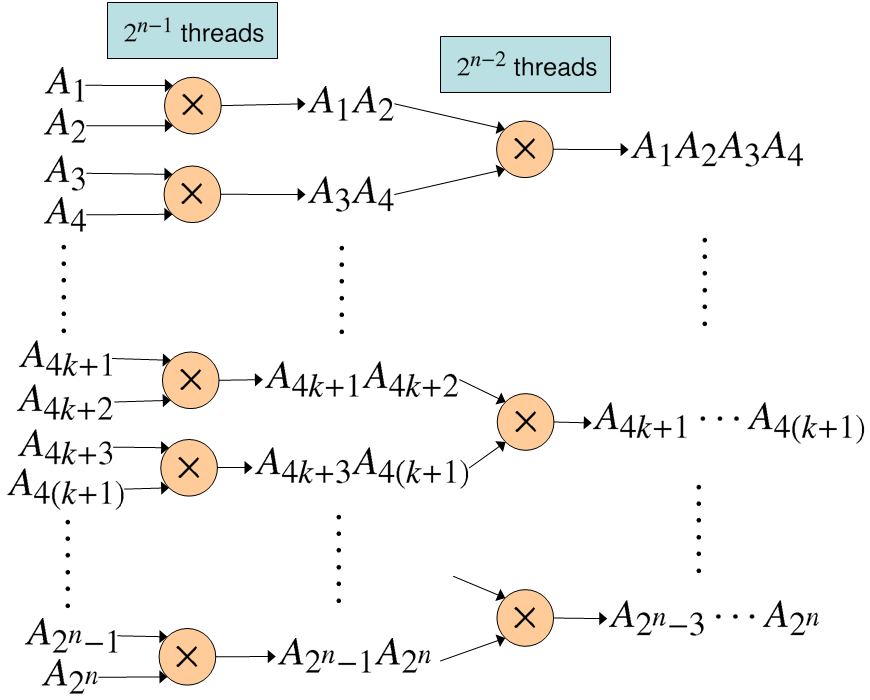}
    \caption{Tree-based multiplication for computing the product $A_1A_2\cdots A_{2^n}$}
    \label{fig:treeMultiply}
  \end{center}
\end{figure}

Direct computation of Equation (\ref{eq:model:eyemotion:prob1}) is intractable. Instead,
we introduce an alternative expression of Equation (\ref{eq:model:eyemotion:prob1}) that
is applicable to stream processing with multi-core processors.
\begin{eqnarray}
  \lefteqn{p(\x(t)|p(S(t)))}\nonumber\\
  &=& \int_{-\infty}^{\infty}\frac{p(s(t,\x(t))=s)}{P(s(t,\x(t))\le s)}
      \prod_{\wt{\x}\in I}P(s(t,\wt{\x})\le s)ds.
      \label{eq:eyemotion:bottomup:1}
\end{eqnarray}
The latter part of Eq. (\ref{eq:eyemotion:bottomup:1}) does not depend on the
position $\x(t)$, which implies that it can be calculated in advance for every $s$. This
calculation can be executed in $\cO(\log|I|)$ time through a tree-based multiplication
and parallelization at each level (cf. Fig. \ref{fig:treeMultiply}). Also, the former
part of Eq. (\ref{eq:eyemotion:bottomup:1}) can be calculated independently for each
position $\x(t)$. Therefore, once the calculation of the latter part has finished, Eq.
(\ref{eq:eyemotion:bottomup:1}) can be calculated in $\cO(\log|S|)$ time with a
combination of tree-based addition and pixel-wise parallelization, where $|S|$ stands for
the resolution of the integral in Eq. (\ref{eq:eyemotion:bottomup:1}).

\subsubsection{Re-sampling}
\label{sec:model:eyemotion:resample}

Finally, re-sampling is performed to eliminate samples with low importance weights and
multiply samples with high importance weights. This step enables us to avoid
``degeneracy'' problem, namely, to avoid the situation where all but one of the
importance weights are close to zero. Although the effective number of samples
\cite{beyondTheKalmanFilter} is frequently used as a criterion for re-sampling, we execute
re-sampling at regular time intervals.

\medskip
\begin{hosoku}\label{rem:particle}
We have to note that the whole procedure which includes the propagation, updating and
re-sampling steps for estimating eye focusing density maps is equivalent to a particle
filter with MCMC sampling since the the PDFs used in the propagation and update steps
are mutually independent with each other. \textit{(Remark\ \ref{rem:particle} ends.)}
\end{hosoku}

\section{Model parameter estimation}
\label{sec:parameter}

\begin{figure}[t]
  \begin{center}
    \includegraphics[width=0.985\hsize,keepaspectratio]{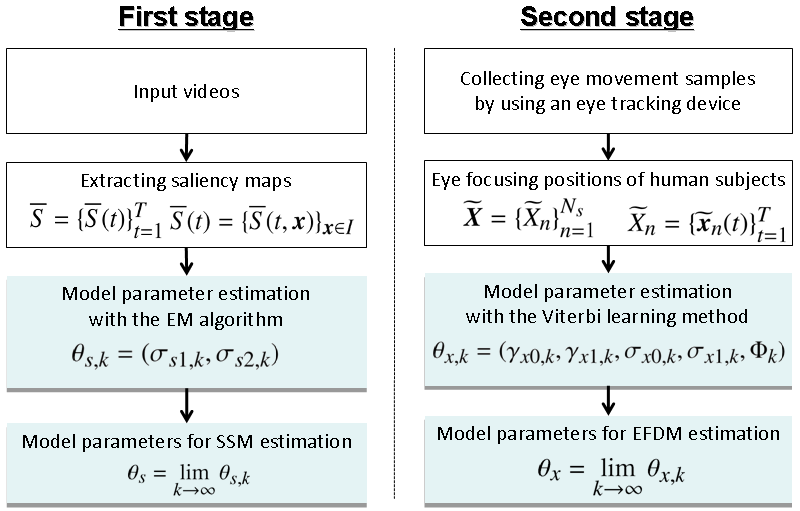}
    \caption{Block diagram of our model parameter estimation}
    \label{fig:paramEstimation}
  \end{center}
\end{figure}

This section focuses on the problem of estimating maximum likelihood (ML) model
parameters. Fig. \ref{fig:paramEstimation} shows the block diagram of our model parameter
estimation. We can automatically estimate almost all the model parameters in advance by
utilizing saliency maps calculated from the input video and eye focusing positions
obtained by some eye tracking devices as observations. Simultaneous estimation of all ML
parameters can be optimal but impractical due to the substantial calculation cost.
Therefore, we separate our parameter estimation into two independent stages. The first stage
derives parameters for computing stochastic saliency maps, and the second stage for
estimating eye focusing points.

\subsection{Parameters for stochastic saliency maps}
\label{sec:parameter:1}

The first stage derives parameters for computing stochastic saliency maps. Here, we
introduce the EM algorithm. In this case, the observations are the saliency maps
$\ov{S}=\ov{S}(1:T)$ and the hidden variables are the stochastic saliency maps
$S=S(1:T)$. Remember that $T$ is the duration of the video. The EM algorithm for
estimating $\theta_s=(\sigma_{s1},\sigma_{s2})$ is as follows:

\medskip\noindent
\ul{$(k+1)$-th E step}

The E step updates the PDF $p(S|\ov{S};\theta_{s,k})$ of the stochastic saliency maps
$S$ given the saliency maps $\ov{S}$ with the previously estimated parameter
$\theta_{s,k}=(\sigma_{s1},\sigma_{s2})$ by using Kalman smoother. In detail, the
objective is to recursively compute the mean $\wh{s}(t|T)$ and standard deviation
$\sigma_s(t|T)$ of the stochastic saliency $s(t)$ at time $t=1,2,\cdots,T$, where all
the saliency maps $\ov{S}$ are used as observations. Note that the position $\y$ is
omitted for simplicity.

Suppose that the PDF of the stochastic saliency at time $t+1$ is given by the following
Gaussian PDF:
\begin{eqnarray*}
  p(s(t+1)|\ov{S};\theta_{s,k})
  &=& \cN(\wh{s}_k(t+1|T),\sigma_{s,k}(t+1|T)).
\end{eqnarray*}
Then, the PDF of the stochastic saliency at time $t$ is obtained by the following
recurrence relation:
\begin{eqnarray*}
  p(s(t)|\ov{S};\theta_{s,k}) &=& \cN(\wh{s}_k(t|T),\sigma_{s,k}(t|T)),
\end{eqnarray*}
where
\begin{eqnarray*}
  \wh{s}_k(t|T)
  &=& \frac{\sigma_{sq,k}^2(t|t)}{\sigma_{s,k}^2(t|t)}\wh{s}_k(t|t)+
      \frac{\sigma_{sq,k}^2(t|t)}{\sigma_{s2,k}^2}\wh{s}_k(t+1|T),\\
  \sigma_{s,k}^2(t|T)
  &=& \sigma_{sq,k}^2(t|t)+\left(\frac{\sigma_{sq,k}^2(t|t)}{\sigma_{s2,k}^2}\right)^2
      \sigma_{s,k}^2(t+1|T),\\
  \sigma_{sq,k}^2(t|t)
  &=& \frac{\sigma_{s2,k}^2\sigma_{s,k}^2(t|t)}{\sigma_{s2,k}^2+\sigma_{s,k}^2(t|t)},
\end{eqnarray*}
and $\wh{s}_k(t|t)$ and $\sigma_{s,k}^2(t|t)$ can be obtained by Eqs.
(\ref{eq:model:stochastic:kalman1})(\ref{eq:model:stochastic:kalman2}) with the
parameter $\theta_{s,k}$.

\medskip\noindent
\ul{$(k+1)$-th M step}

The M step updates the parameter $\theta_s$ to maximize the expected log likelihood of
the PDF $p(\ov{S},S;\theta_s)$. We can derive a new parameter $\theta_{s,k+1}$ from
the result of the E step by taking the derivatives of the log likelihood in terms of
$\theta_s$ and setting to 0.
\begin{eqnarray*}
  \lefteqn{\sigma_{s1,k+1}^2 = \frac{1}{T}\sum_{t=1}^{T}
      \left\{(\ov{s}(t)-\wh{s}_k(t|T))^2+\sigma_{s,k}^2(t|T)\right\}}\\
  \lefteqn{\sigma_{s2,k+1}^2 = \frac{1}{T-1}\sum_{t=2}^{T}
           \Bigl[(\wh{s}_k(t|T)-\wh{s}_k(t-1|T))^2}\\
  & & \hspace{-5mm}\left.+\sigma_{s,k}^2(t-1|T)
      +\frac{\sigma_{s2,k}^2-\sigma_{s,k}^2(t-1|t-1)}
            {\sigma_{s2,k}^2+\sigma_{s,k}^2(t-1|t-1)}\sigma_{s,k}^2(t|T)\right]
\end{eqnarray*}

\subsection{Parameters for eye focusing positions}
\label{sec:parameter:2}

The second stage derives parameters $\theta_x$ $=$ $(\gamma_{x0}$, $\sigma_{x0}$,
$\gamma_{x1}$, $\sigma_{x1}$, $\Phi)$ for computing eye focusing positions. The
observations are the sequence of eye focusing positions $\X=\x(1:T)$ obtained from some
eye tracking devices, and the hidden states are the eye movement patterns $U=u(1:T)$. In
this section, we introduce an alternative notation of eye movement patterns as
$u(t)=(u(t)_0,u(t)_1)^T$, which is a 2-dimensional binary vector such that $(1,0)^T$
denotes the passive state, and $(0,1)^T$ represents the active state.

We take a Viterbi learning approach for its quick convergence. It recursively updates the
eye movement patterns $U=u(1:T)$ and the ML parameter set $\theta_{x}$ to maximize the
posterior $p(U|\X;\theta_{x})$.

\medskip\noindent
\ul{Initializing eye movement patterns}

We have to start with determining an initial sequence $U_0=u_0(1:T)$ of eye
movement patterns. We introduce the following decision rule:
\begin{eqnarray*}
  u_0(t) &=& \left\{
  \begin{array}{ll}
    0 & \quad\mbox{if }\|\x(t)-\x(t-1)\|\le\kappa_x\\
    1 & \quad\mbox{if }\|\x(t)-\x(t-1)\| > \kappa_x
  \end{array}\right.,
\end{eqnarray*}
where $\kappa_x$ is a given threshold.

\medskip\noindent
\ul{The $(k+1)$-th step for updating hidden variables}

This step updates the sequence $U$ of eye movement patterns to maximize the
posterior density $p(U|\X;\theta_{x,k})$ given the parameter set $\theta_{x,k}$
obtained in the previous step.
\begin{eqnarray*}
  U_{k+1} &\leftarrow& \arg\max_U p(U|\X;\theta_{x,k}).
\end{eqnarray*}
Viterbi algorithm (e.g. \cite{Viterbi,Viterbi:Rabiner}) can derive the ML sequence
$U_{k+1}$ of eye movement patterns.

\medskip\noindent
\ul{The $(k+1)$-th step for updating the parameter set}

This step updates the parameter set $\theta_x$ to maximize the posterior density
$p(U_{k+1}|\X;\theta_x)$.
\begin{eqnarray*}
  \theta_{x,k+1} &=& \arg\max_{\theta_x}p(U_{k+1}|\X;\theta_x).
\end{eqnarray*}
Taking the derivative of the log likelihood in terms of $\theta_x$, we obtain
\begin{eqnarray*}
  \lefteqn{\gamma_{xi,k+1}}\\
  &=& \frac{\sum_{t=2}^{T} \|\x(t)-\x(t-1)\| u_{k+1}(t)_i}
           {\sum_{t=2}^{T} u_{k+1}(t)_i},\\
  \lefteqn{\sigma_{xi,k+1}^2}\\
  &=& \frac{\sum_{t=2}^{T} (\|\x(t)-\x(t-1)\|-\gamma_{xi,k})^2 u_{k+1}(t)_i}
           {2\sum_{t=2}^{T} u_{k+1}(t)_i},\\
  \lefteqn{{\phi_{(i,j)}}_{,k+1}}\\
  &=& \frac{\sum_{t=2}^{T} u_{k+1}(t)_i u_{k+1}(t-1)_j}
           {\sum_{t=2}^{T} u_{k+1}(t-1)_j}.
\end{eqnarray*}

\section{Evaluation}
\label{sec:eval}

\subsection{Evaluation conditions}
\label{sec:eval:condition}

For the accuracy evaluation, we used CRCNS eye-1 database created by University of South
California. This database includes 100 video clips (MPEG-1, $640\times 480$ pixels,
30fps) and eye traces when showing these video clips to 8 human subjects (4-6 available
eye traces for each video clip, 240fps). Other details for the database can be found in
\textit{\url{https://crcns.org/files/data/eye-1/crcns-eye1-summary.pdf}}. In this
evaluation, we used 50 video clips (about 25 minutes in total) called ``original
experiment'' and associated eye traces.

Model parameters were derived in advance with the learning algorithm presented in
Section \ref{sec:parameter}. In this time, we used 5-fold cross validation so that
40 video clips and associated eye traces were used as the training data for evaluating
the remaining data (10 videos and associated eye traces).

All the algorithms were implemented with a standard C++ platform and NVIDIA CUDA, and the
evaluation were carried out on a standard PC with a graphics
processor unit (GPU). The detailed information for the platform used in this evaluation
is listed in Table \ref{table:platform}.

\begin{table}[t]
  \begin{center}
    \caption{Platform used in the evaluation}
    \label{table:platform}
    \begin{tabular}{|l||r|}\hline
      OS & Windows Vista Ultimate\\\hline
      Development & Microsoft Visual C++ 2008\\
      platform & OpenCV 1.1pre \& NVIDIA CUDA 2.2\\\hline
      Optimization & disabled\\\hline
      CPU & Intel Core2 Quad Q6600 (2.40GHz)\\\hline
      RAM & 4.0GB\\\hline
      GPU & NVIDIA GeForce GTX275 $\times 2$ SLI\\\hline
    \end{tabular}
  \end{center}
\end{table}

\subsection{Evaluation metric}
\label{sec:eval:metric}

As a metric to quantify how well a model predicts the actual human eye focusing
positions, we used the normalized scan-path saliency (NSS) used in the previous work
\cite{beyondBottomUp}. Let $R_j(t)$ be a set of all pixels in a circular region centered
on the eye focusing position of test subject $j$ with a radius of 30 pixels. Then, the
NSS value at time $t$ is defined as
\begin{eqnarray*}
  NSS(t) = \frac{1}{N_s}\sum_{j=1}^{N_s}\frac{1}{\sigma(p(\x))}
           \left\{\max_{\x(t)\in R_j(t)}p(\x(t))-\ov{p}(\x)\right\},
\end{eqnarray*}
where $N_s$ is the total number of subjects, $\ov{p}(\x)$ and $\sigma(p(\x))$ are the
mean and the variance of the pixel values of the model's output, respectively. $NSS(t)=1$
indicates that the subjects' eye positions fall in a region whose predicted density is
one standard deviation above average. Meanwhile, $NSS(t)\le 0$ indicates that the model
performs no better than picking a random position on the map. 

\subsection{Results}
\label{sec:eval:results}

We compared our proposed method with 3 existing computational models: 1) a simple control
measuring local pixel variance (denoted ``variance'') \cite{DetectSurprise:itti}, 2) a
saliency map (denoted ``CIOFM'') \cite{Saliency:itti}, and 3) Bayesian surprise (denoted
``surprise'') \cite{DetectSurprise:itti}. All the outputs emitted from the above existing
models are included in CRCNS eye-1 database, and therefore we directly utilized them for
the evaluation.

\begin{figure}[t]
  \begin{center}
    \includegraphics[width=0.985\hsize,keepaspectratio]{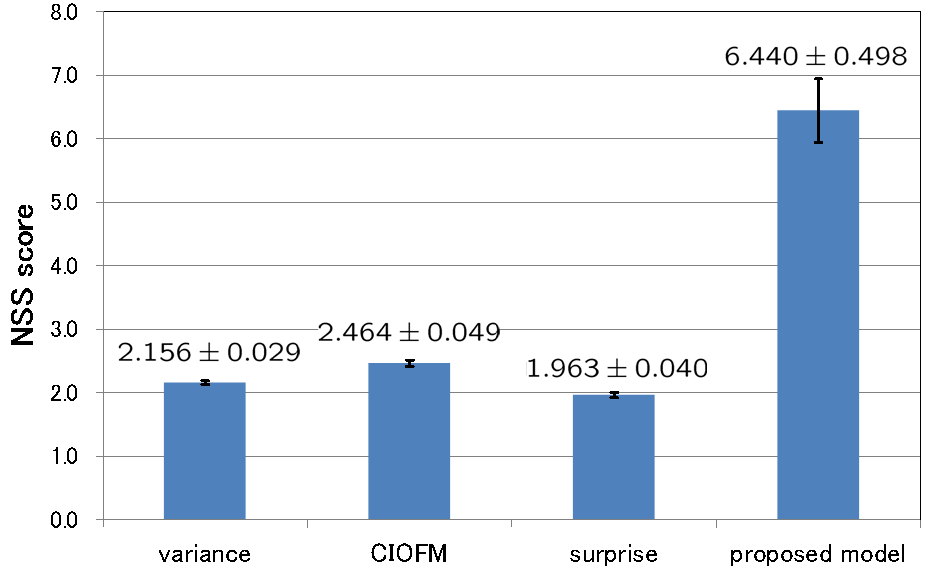}
    \caption{Average NSS score for all the video clips}
    \label{fig:result_NSS}
  \end{center}
\end{figure}

\begin{figure*}[t]
  \begin{center}
    \includegraphics[width=0.985\hsize,keepaspectratio]{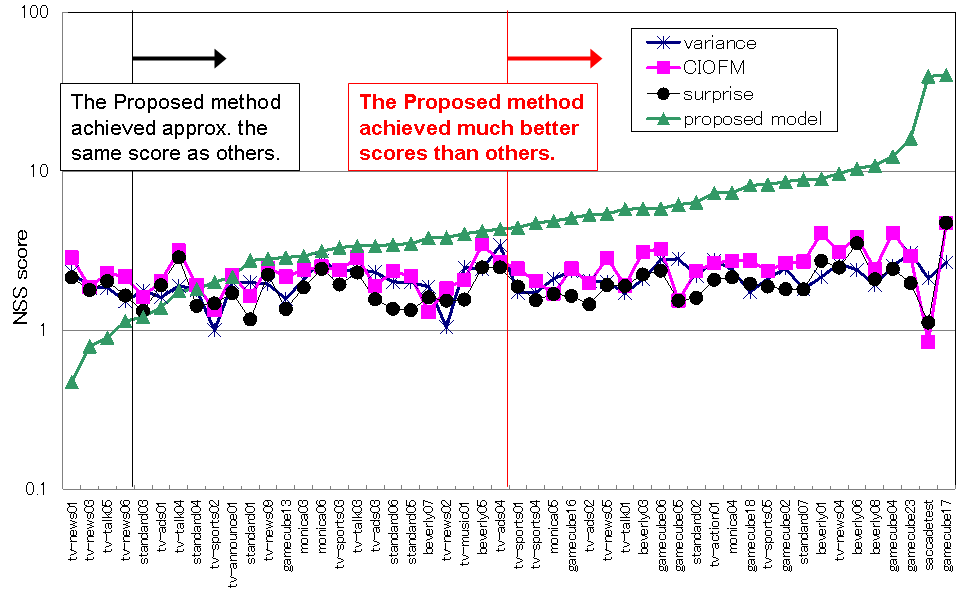}
    \caption{
      Average NSS score for each video clip, where the vertical axis uses a log scale
      and video clips on the horizontal axis are sorted in the ascending order of NSS
      scores.
    }
    \label{fig:result_NSS_each}
  \end{center}
\end{figure*}

Fig. \ref{fig:result_NSS} shows the model accuracy measured by the average NSS score with
standard errors for all the video clips, and Fig. \ref{fig:result_NSS_each} details the
average NSS score for each video clip. The order of video clips is sorted beforehand to
keep the visibility. The result shown in Fig. \ref{fig:result_NSS} indicates that the our
new method achieved significantly better scores than all 3 existing methods, which
implies that our proposed method can estimate human visual attention with high accuracy.
Also, the result shown in Fig. \ref{fig:result_NSS_each} indicates that our proposed
method marked almost the same as or much better than all the existing methods for most of
the video clips.

\begin{figure*}[t]
  \begin{center}
    \includegraphics[width=0.985\hsize,keepaspectratio]{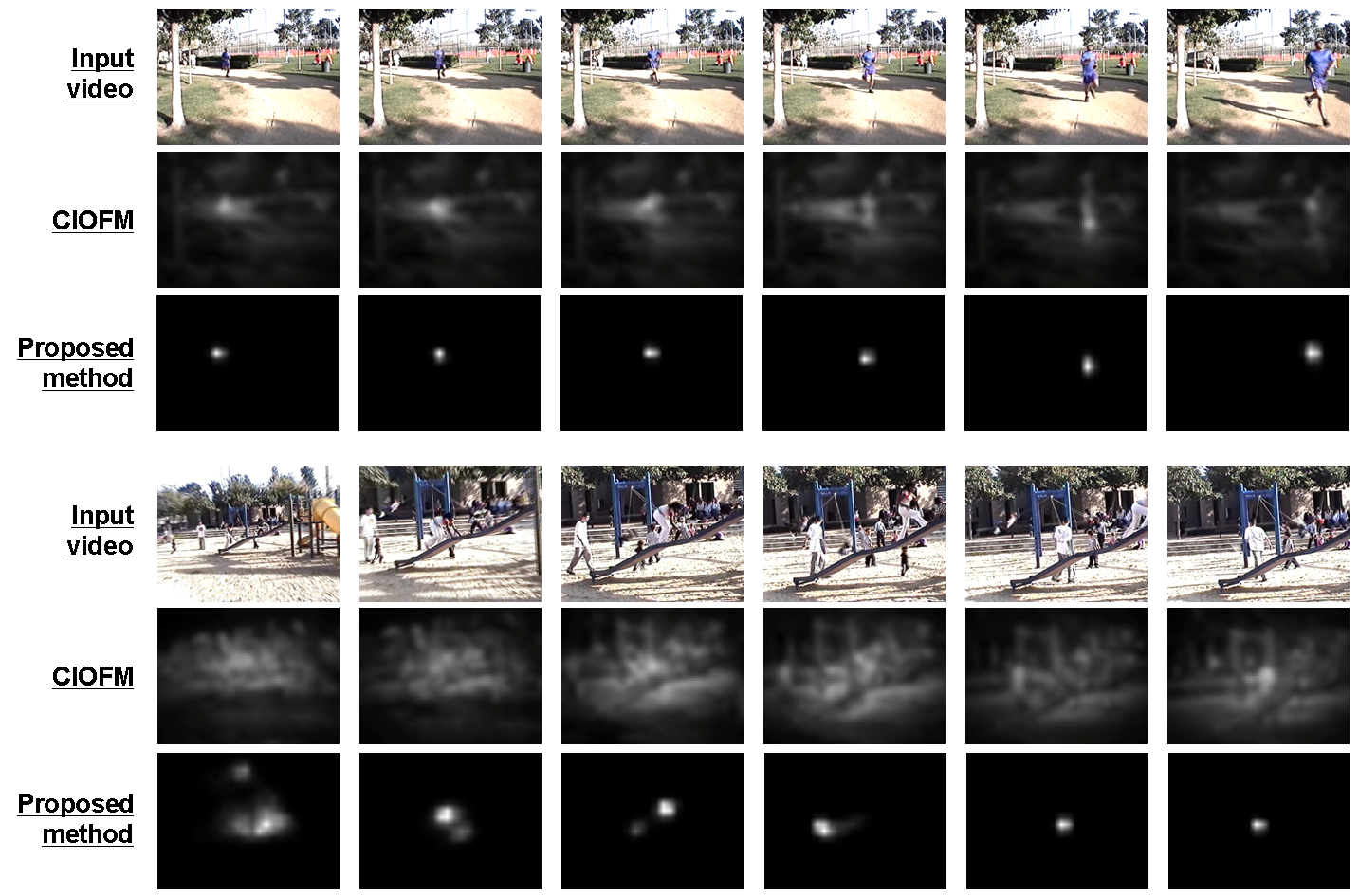}
    \caption{Snapshots of model outputs}
    \label{fig:snapshot}
  \end{center}
\end{figure*}

Fig. \ref{fig:snapshot} shows snapshots of outputs from Itti model (the second and fifth
rows) and our proposed method (the third and sixth rows). It illustrates that outputs
from Itti model included several large salient regions. On the other hand, outputs from
our proposed method included only a few small eye focusing areas. This implies that our
new method picked up probable eye focusing areas accurately.

\begin{figure}[t]
  \begin{center}
    \includegraphics[width=0.985\hsize,keepaspectratio]{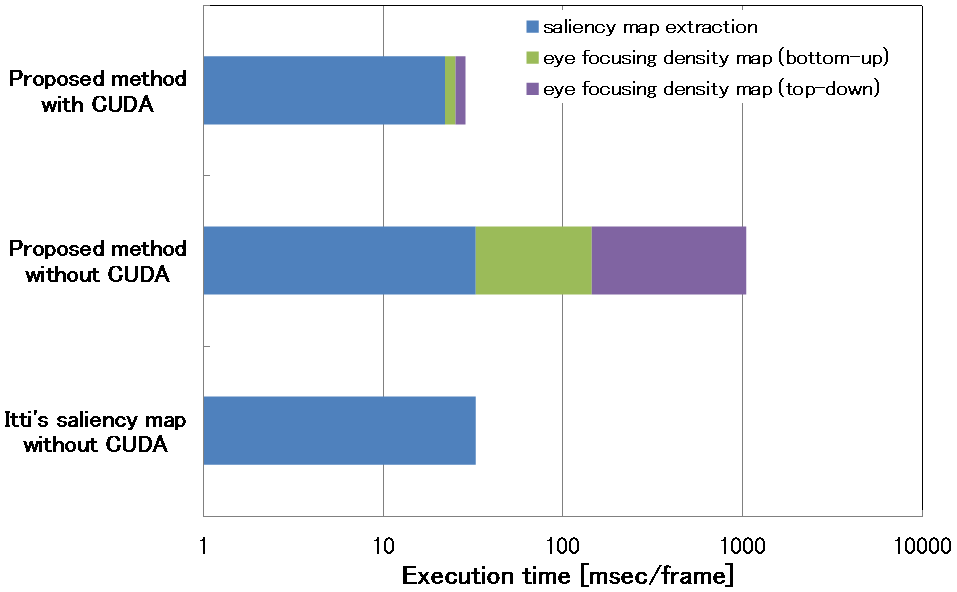}
    \caption{Total execution time [msec/frame] (log-scaled)}
    \label{fig:result_time}
  \end{center}
\end{figure}

Fig. \ref{fig:result_time} shows the total execution time of 1) calculating Itti's
saliency map without CUDA, 2) the proposed method without CUDA, and 3) the proposed
method with CUDA. The result indicates that the proposed method has achieved near
real-time estimation (40-50 msec/frame), and almost the same processing time as the one
for Itti's model.

\section{Conclusion}
\label{sec:conclude}

We have presented the first stochastic model of human visual attention based on a dynamic
Bayesian framework. Unlike many existing methods, we predict the likelihood of human-%
attended regions on a video based on two criteria: 1) The probability of having the
maximum saliency response at a given region evaluated based on the signal detection
theory, and 2) the probability of matching the eye movement projection based on the 
predicted state. Experiments have revealed that our model offers a better eye-gazing
prediction against previous deterministic models. To enhance our current model, future
work may include determination of initial parameters close to the global optima when
estimating model parameters, unified approach to estimate all the model parameters,
a better density model of eye movements, a better integration of the bottom-up and the
top-down information, a better saliency model for extracting (deterministic) saliency
maps, and integration of the proposed method into some applications such as driving
assistance, active vision and video retrieval.

\section*{Acknowledgment}

The authors thank Prof. Laurent Itti of University of South California, Prof. Minho Lee
of Kyungpook National University, Dr. Hirokazu Kameoka and Dr. Eisaku Maeda of NTT
Communication Science Laboratories for their
valuable discussions and helpful comments, which led to improvements of this work. The
second and fourth authors contributed to this work during their internship at NTT
Communication Science Laboratories. The authors also thank Dr. Yoshinobu Tonomura
(currently Ryukoku University), Dr.
Hiromi Nakaiwa, Dr. Naonori Ueda, Dr. Hiroshi Sawada, Dr. Shoji Makino (currently Tsukuba
University) and Dr. Kenji Nakazawa (currently NTT Advance Technologies Inc.) of NTT
Communication Science Laboratories for their help to the internship.

\bibliographystyle{bib/IEEEtran}
\bibliography{bib/IEEEabrv,bib/defs,bib/retrieval,bib/saliency,bib/statistics}

\begin{thebibliography}{10}
\providecommand{\url}[1]{#1}
\csname url@samestyle\endcsname
\providecommand{\newblock}{\relax}
\providecommand{\bibinfo}[2]{#2}
\providecommand{\BIBentrySTDinterwordspacing}{\spaceskip=0pt\relax}
\providecommand{\BIBentryALTinterwordstretchfactor}{4}
\providecommand{\BIBentryALTinterwordspacing}{\spaceskip=\fontdimen2\font plus
\BIBentryALTinterwordstretchfactor\fontdimen3\font minus
  \fontdimen4\font\relax}
\providecommand{\BIBforeignlanguage}[2]{{%
\expandafter\ifx\csname l@#1\endcsname\relax
\typeout{** WARNING: IEEEtran.bst: No hyphenation pattern has been}%
\typeout{** loaded for the language `#1'. Using the pattern for}%
\typeout{** the default language instead.}%
\else
\language=\csname l@#1\endcsname
\fi
#2}}
\providecommand{\BIBdecl}{\relax}
\BIBdecl

\bibitem{roboticsSaliency:asada}
M.~Hikita, S.~Fuke, M.~Ogino, T.~Minato, and M.~Asada, ``Visual attention by
  saliency leads cross-modal body representation,'' in \emph{Development and
  Learning, 2008. ICDL 2008. 7th IEEE International Conference on}, Aug. 2008,
  pp. 157--162.

\bibitem{medicalImagingSaliency}
X.-P. Hu, L.~Dempere-Marco, and G.-Z. Yang, ``Hot spot detection based on
  feature space representation of visual search,'' \emph{Medical Imaging, IEEE
  Transactions on}, vol.~22, no.~9, pp. 1152--1162, Sept. 2003.

\bibitem{intelligentCarsSaliency}
P.~Santana, M.~Guedes, L.~Correia, and J.~Barata, ``Saliency-based obstacle
  detection and ground-plane estimation for off-road vehicles,'' in
  \emph{ICVS}, 2009, pp. 275--284.

\bibitem{nonparaBayesAttention}
G.~Boccignone, ``Nonparametric bayesian attentive video analysis,'' in
  \emph{Proc. International Conference on Pattern Recognition (ICPR)}, Dec.
  2008, pp. 1--4.

\bibitem{foveatedShotDetection:boccignone}
G.~Boccignone, A.~Chianese, V.~Moscato, and A.~Picariello, ``Foveated shot
  detection for video segmentation,'' \emph{Circuits and Systems for Video
  Technology, IEEE Transactions on}, vol.~15, no.~3, pp. 365--377, March 2005.

\bibitem{salientRegionExtraction:fukuchi_ICME}
K.~Fukuchi, K.~Miyazato, A.~Kimura, S.~Takagi, and J.~Yamato, ``Saliency-based
  video segmentation with graph cuts and sequentially updated priors,'' in
  \emph{Proc. International Conference on Multimedia and Expo (ICME)}, June
  2009.

\bibitem{nearIdenticalShotDetection:zisserman}
O.~Chum, J.~Philbin, M.~Isard, and A.~Zisserman, ``Scalable near identical
  image and shot detection,'' in \emph{CIVR '07: Proceedings of the 6th ACM
  international conference on Image and video retrieval}.\hskip 1em plus 0.5em
  minus 0.4em\relax New York, NY, USA: ACM, 2007, pp. 549--556.

\bibitem{visualAttention:KochUllman}
C.~Koch and S.~Ullman, ``Shifts in selective visual attention: Towards the
  underlying neural circuitry,'' \emph{Human Neurobiology}, vol.~4, pp.
  219--227, 1985.

\bibitem{FeatureIntegrationTheory:Treisman}
A.~Treisman and G.~Gelade, ``A feature-integration theory of attention,''
  \emph{Cognitive Psychology}, vol.~12, pp. 97--136, 1980.

\bibitem{Saliency:itti}
L.~Itti, C.~Koch, and E.~Niebur, ``A model of saliency-based visual attention
  for rapid scene analysis,'' \emph{{IEEE} Trans. Pattern Anal. Mach. Intell.},
  vol.~20, no.~11, pp. 1254--1259, November 1998.

\bibitem{VisualAttention:Privitera}
C.~M. Privitera and L.~W. Stark, ``Algorithms for defining visual
  regions-of-interest: Comparison with eye fixations,'' \emph{{IEEE} Trans.
  Pattern Anal. Mach. Intell.}, vol.~22, no.~9, pp. 970--982, 2000.

\bibitem{GeneratingEyeFixations:gu}
E.~Gu, J.~Wang, and N.~Badler, ``Generating sequence of eye fixations using
  decision-theoretic attention model,'' in \emph{Proc. Conference on Computer
  Vision and Pattern Recognition (CVPR)}, June 2005, pp. 92--99.

\bibitem{VOCUS}
\BIBentryALTinterwordspacing
S.~Frintrop, \emph{Vocus: a Visual Attention System for Object Detection And
  Goal-directed Search (Lecture Notes in Computer Science)}.\hskip 1em plus
  0.5em minus 0.4em\relax Springer-Verlag New York Inc (C), 3 2006. [Online].
  Available: \url{http://amazon.co.jp/o/ASIN/3540327592/}
\BIBentrySTDinterwordspacing

\bibitem{saliency:minho2}
S.~Jeong, S.~Ban, and M.~Lee, ``Stereo saliency map considering affective
  factors and selective motion analysis in a dynamic environment,''
  \emph{Neural Networks}, vol.~21, pp. 1420--1430, October 2008.

\bibitem{decisionTheorySaliency:gao}
D.~Gao and N.~Vasconcelos, ``Decision-theoretic saliency: Computational
  principles, biological plausibility, and implications for neurophysiology and
  psychophysics,'' \emph{Neural Computation}, vol.~21, no.~1, pp. 239--271,
  January 2009.

\bibitem{DetectSurprise:itti}
L.~Itti and P.~Baldi, ``A principled approach to detecting surprising events in
  video,'' in \emph{Proc. Conference on Computer Vision and Pattern Recognition
  (CVPR)}, June 2005, pp. 631--637.

\bibitem{videoSaliency:clement}
C.~Leung, A.~Kimura, T.~Takeuchi, and K.~Kashino, ``A computational model of
  saliency depletion/recovery phenomena for the salient region extraction of
  videos,'' in \emph{Proc. International Conference on Multimedia and Expo
  (ICME)}, July 2007, pp. 300--303.

\bibitem{saliency:minho1}
S.~Ban, I.~Lee, and M.~Lee, ``Dynamic visual selective attention model,''
  \emph{Neurocomputing}, vol.~71, pp. 853--856, March 2007.

\bibitem{SignalDetectionTheory:Eckstein}
M.~P. Eckstein, J.~P. Thomas, J.~Palmer, and S.~S. Shimozaki, ``A signal
  detection model predicts effects of set size on visual search accuracy for
  feature, conjunction, triple conjunction and disjunction displays,''
  \emph{Perception and Psychophysics}, vol.~62, pp. 425--451, 2000.

\bibitem{SignalDetectionTheory:Verghese}
P.~Verghese, ``Visual search and attention: A signal detection theory
  approach,'' \emph{Neuron}, vol.~31, pp. 525--535, August 2001.

\bibitem{objectsAttention:Scholl2002}
\BIBentryALTinterwordspacing
B.~J. Scholl, Ed., \emph{Objects and Attention (Cognition Special
  Issue)}.\hskip 1em plus 0.5em minus 0.4em\relax The MIT Press, 8 2002.
  [Online]. Available: \url{http://amazon.co.jp/o/ASIN/0262692805/}
\BIBentrySTDinterwordspacing

\bibitem{AvrahamLindenbaumPaper}
T.~Avraham and M.~Lindenbaum, ``Esaliency (extended saliency): Meaningful
  attention using stochastic image modeling,'' \emph{IEEE Transactions on
  Pattern Analysis and Machine Intelligence}, vol.~32, pp. 693--708, 2009.

\bibitem{spaciotemporalSaliency}
V.~Mahadevan and N.~Vasconcelos, ``Spatiotemporal saliency in dynamic scenes,''
  \emph{IEEE Transactions on Pattern Analysis and Machine Intelligence},
  vol.~32, pp. 171--177, 2009.

\bibitem{beyondBottomUp}
R.~J. Peters and L.~Itti, ``Beyond bottom-up: Incorporating task-dependent
  influences into a computational model of spatial attention,'' in \emph{Proc.
  Conference on Computer Vision and Pattern Recognition (CVPR)}, June 2007, pp.
  1--8.

\bibitem{adaptiveKalman:myers}
K.~Myers and B.~Taplay, ``Adaptive sequential estimation with unknown noise
  statistics,'' \emph{{IEEE} Trans. Autom. Control}, vol.~21, no.~4, pp.
  520--523, August 1976.

\bibitem{adaptiveKalman:leathrum}
J.~Leathrum, ``On sequential estimation of state noise variances,''
  \emph{{IEEE} Trans. Autom. Control}, vol.~26, no.~3, pp. 745--746, June 1981.

\bibitem{Metropolis}
N.~Metropolis, A.~Rosenbluth, M.~Rosenbluth, A.~Teller, and E.~Teller,
  ``Equation of state calculations by fast computing machines,'' \emph{Journal
  of Chemical Physics}, vol.~21, pp. 1087--1092, 1953.

\bibitem{beyondTheKalmanFilter}
B.~Ristic, S.~Arulampalam, and N.~Gordon, \emph{Beyond the {K}alman filter:
  {P}article filters for tracking applications}.\hskip 1em plus 0.5em minus
  0.4em\relax Boston: Artech {H}ouse {P}ublishers, 2004.

\bibitem{Viterbi}
A.~Viterbi, ``Error bounds for comvolutional codes and an asymptotically
  optimum decoding algorithm,'' \emph{{IEEE} Trans. Inf. Theory}, vol.~13,
  no.~2, pp. 260--269, April 1967.

\bibitem{Viterbi:Rabiner}
L.~Rabiner, ``A tutorial on hidden {M}arkov models and selected applications in
  speech recognition,'' \emph{Proc. the IEEE}, vol.~77, no.~2, pp. 257--286,
  February 1989.

\end{thebibliography}

\begin{IEEEbiography}[{\includegraphics[width=1in,height=1.25in,clip,keepaspectratio]%
{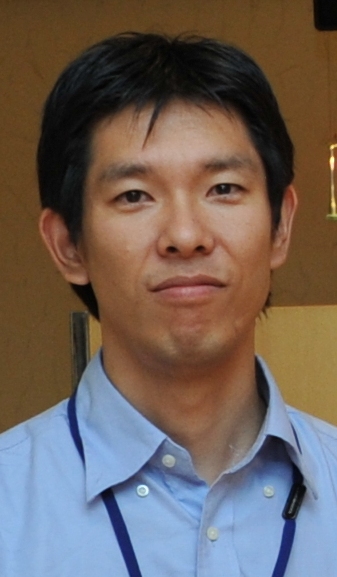}}]{Akisato Kimura}
(M'00-SM'07) received B.E., M.E. and D.E. degrees in Communications and Integrated
Systems from Tokyo Institute of Technology, Japan in 1998, 2000 and 2007, respectively.
Since 2000, he has been with NTT Communication Science Laboratories, Nippon Telegraph and
Telephone Corporation, where he is currently a Research Scientist in Media Information
Laboratory. He has been working on content-based multimedia copy detection, image/video
retrieval, automatic image/video annotation, computational models of human visual
attention and visual scene understanding. His research interests include pattern
recognition, computer vision, image/video processing, human visual perception,
cognitive science, statistical signal processing, machine learning and information
theory.%
\end{IEEEbiography}%
\begin{IEEEbiography}[{\includegraphics[width=1in,height=1.25in,clip,keepaspectratio]%
{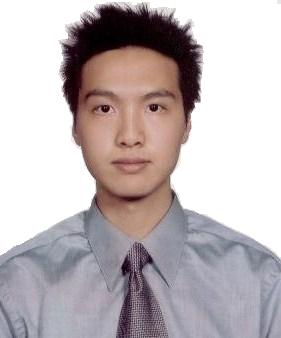}}]{Derek Pang}
(S'04) received a B.A.Sc. degree from Simon Fraser University, Canada in 2009. Since
September 2009, he has been with Department of Electrical Engineering, Stanford
University, where he is a Ph.D student in the Image, Video and Multimedia Systems (IVMS)
group. From June to December 2007, he stayed at Media Information Laboratory, NTT
Communication Science Laboratories, Nippon Telegraph and Telephone Corporation as a
research intern. He received the Student Travel Award in IEEE International Conference on
Multimedia and Expo (ICME2009). His research interests are in the areas of computer
vision, machine learning and multimedia processing and coding.%
\end{IEEEbiography}%
\begin{IEEEbiography}[{\includegraphics[width=1in,height=1.25in,clip,keepaspectratio]%
{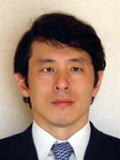}}]{Tatsuto Takeuchi}
received B.A. and M.A. degrees from Kyoto University, Japan and the Ph.D. degree in
experimental psychology from the University of Tokyo, Japan in 1994. He performed
postdoctoral research at the University of California at Berkeley. In 1995, he joined NTT
Basic Research Laboratories (now NTT Communication Science Laboratories), Nippon
Telegraph and Telephone Corporation. He is currently working on basic and applied
research concerning visual motion and color perception. He is a member of the Association
for Research in Vision and Ophthalmology, the Japanese Psychological Association and the
Vision Society of Japan.%
\end{IEEEbiography}%
\begin{IEEEbiography}[{\includegraphics[width=1in,height=1.25in,clip,keepaspectratio]%
{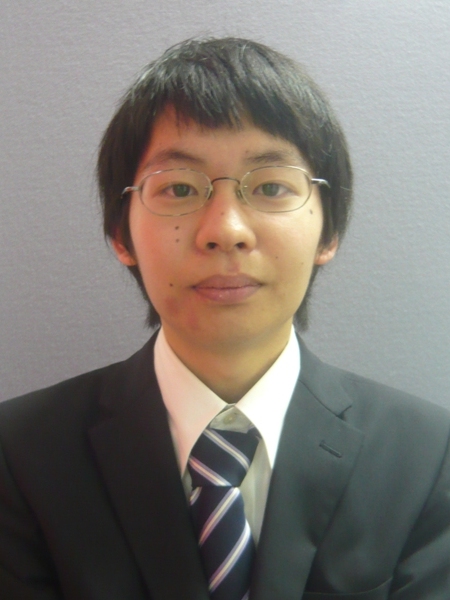}}]{Kouji Miyazato}
has been with Department of Information and Communication Systems Engineering, Okinawa
National College of Technology, Japan, where he is a bachelor student. In September 2008,
he stayed at Media Information Laboratory, NTT Communication Science Laboratories, Nippon
Telegraph and Telephone Corporation as a research intern. During his internship, he was
working on research of stream processing with GPUs. He received the Outstanding
Performance Award in the 6th Japanese Olympiad in Informatics (JOI 2006/2007).%
\end{IEEEbiography}%
\begin{IEEEbiography}[{\includegraphics[width=1in,height=1.25in,clip,keepaspectratio]%
{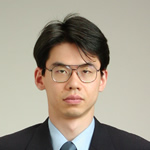}}]{Kunio Kashino}
(S'89-M'95-SM'05) received a Ph.D. degree in electrical engineering from the University
of Tokyo in 1995. In 1995, he joined NTT Corporation, where he is currently a Senior
Research Scientist, Supervisor. He has been working on multimedia information retrieval.
His research interests include acoustic signal processing and Bayesian information
integration.%
\end{IEEEbiography}%
\begin{IEEEbiography}[{\includegraphics[width=1in,height=1.25in,clip,keepaspectratio]%
{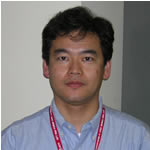}}]{Junji Yamato}
(M'90-SM'05) received B.E. and M.E. degrees from the University of Tokyo, Japan, in 1988
and 1990, respectively. He received an M.S. degree from the MIT in 1998, and a Ph.D.
degree from the University of Tokyo in 2001. He is currently the Executive Manager of
Media Information Laboratory, NTT Communication Science Laboratories. His research
interests include computer vision, gesture recognition and human-robot interaction.%
\end{IEEEbiography}%

\vfill
\pagebreak

\end{document}